\documentclass[10pt,twocolumn,letterpaper]{article}

\usepackage{cvpr}              

\usepackage{graphicx}
\usepackage{amsmath}
\usepackage{amssymb}
\usepackage{booktabs}
\usepackage{hhline}

\usepackage[pagebackref,breaklinks,colorlinks]{hyperref}

\usepackage[capitalize]{cleveref}
\crefname{section}{Sec.}{Secs.}
\Crefname{section}{Section}{Sections}
\Crefname{table}{Table}{Tables}
\crefname{table}{Tab.}{Tabs.}

\def\cvprPaperID{*****} 
\def\confName{CVPR}
\def\confYear{2024}

\begin{document}

\title{MUSTAN: Multi-scale Temporal Context as Attention for Robust Video Foreground Segmentation}


\author{Praveen Kumar Pokala\\
AI-CoE, Jio Platforms Limited (JPL), India\\
{\tt\small praveenkumar.pokala@gmail.com}
\and
 Jaya Sai Kiran Patibandla\\
 AI-CoE, Jio Platforms Limited (JPL), India\\
{\tt\small jaya.patibandla@ril.com}
\and
 Naveen Kumar Pandey\\
 AI-CoE, Jio Platforms Limited (JPL), India\\
{\tt\small  naveen2.pandey@ril.com}
\and
Balakrishna Reddy Pailla\\
AI-CoE, Jio Platforms Limited (JPL), India\\
{\tt\small  balakrishna.pailla@ril.com}
}

\maketitle

\begin{abstract}
   Video foreground segmentation (VFS) is an important computer vision task wherein one aims to segment the objects under motion from the background. Most of the current methods are image-based, i.e., rely only on spatial cues while ignoring motion cues. Therefore, they tend to overfit the training data and don't generalize well to out-of-domain (OOD) distribution. To solve the above problem, prior works exploited several cues such as optical flow, background subtraction mask, etc. However, having a video data with annotations like optical flow is a challenging task. In this paper, we utilize the temporal information and the spatial cues from the video data to improve OOD performance. However, the challenge lies in how we model the temporal information given the video data in an interpretable way creates a very noticeable difference. We therefore devise a strategy that integrates the temporal context of the video in the development of VFS. Our approach give rise to deep learning architectures, namely MUSTAN1 and MUSTAN2 and they are based on the idea of multi-scale temporal context as an attention, i.e., aids our models to learn better representations that are beneficial for VFS.  Further, we introduce a new video dataset, namely Indoor Surveillance Dataset (ISD) for VFS. It has multiple annotations on a frame level such as foreground binary mask, depth map, and instance semantic annotations. Therefore, ISD can benefit other computer vision tasks. We validate the efficacy of our architectures and compare the performance with baselines. We demonstrate that proposed methods significantly outperform the benchmark methods on OOD. In addition, the performance of MUSTAN2 is significantly improved on certain video categories on OOD data due to ISD.

\end{abstract}

\begin{figure*}[!t]
    \centering
    \includegraphics[scale=0.22]{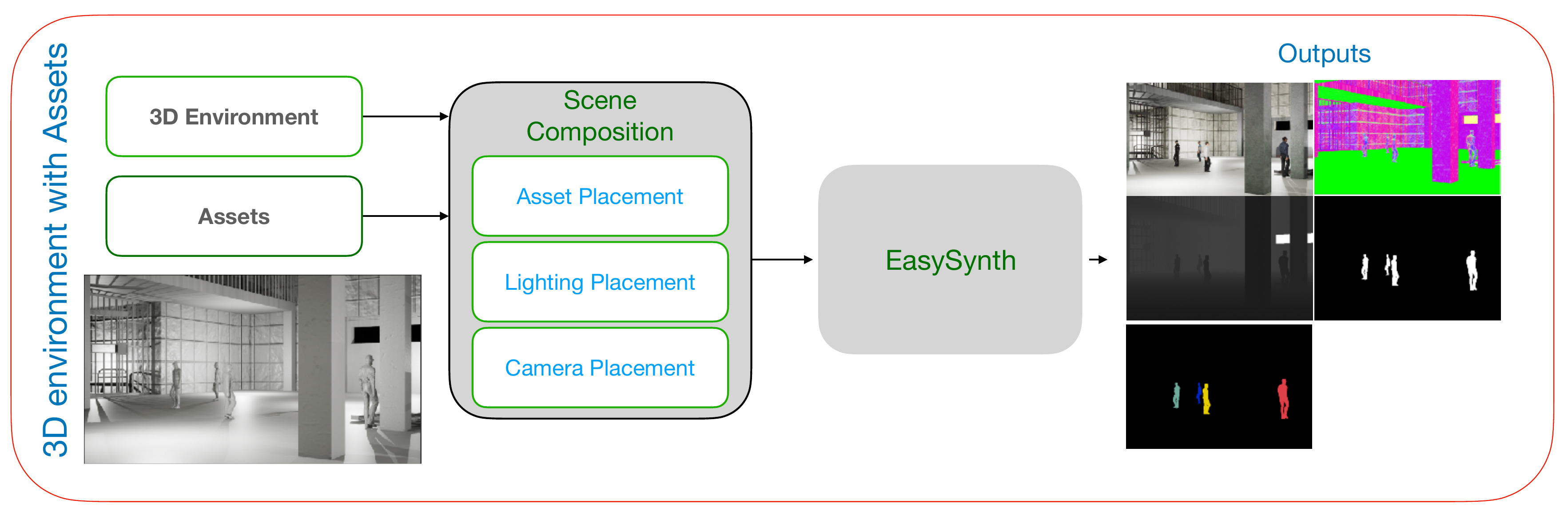}
    \caption{\centering {\bf ISD} Pipeline that takes 3D environment along with assets like humans as input and outputs synthetic RGB image with various annotations such as binary foreground mask, normal map, depth map, and instance semantic map. }
    \label{dgp_}
\end{figure*}
\begin{figure*}[!t]
    \centering
    \includegraphics[scale=0.25]{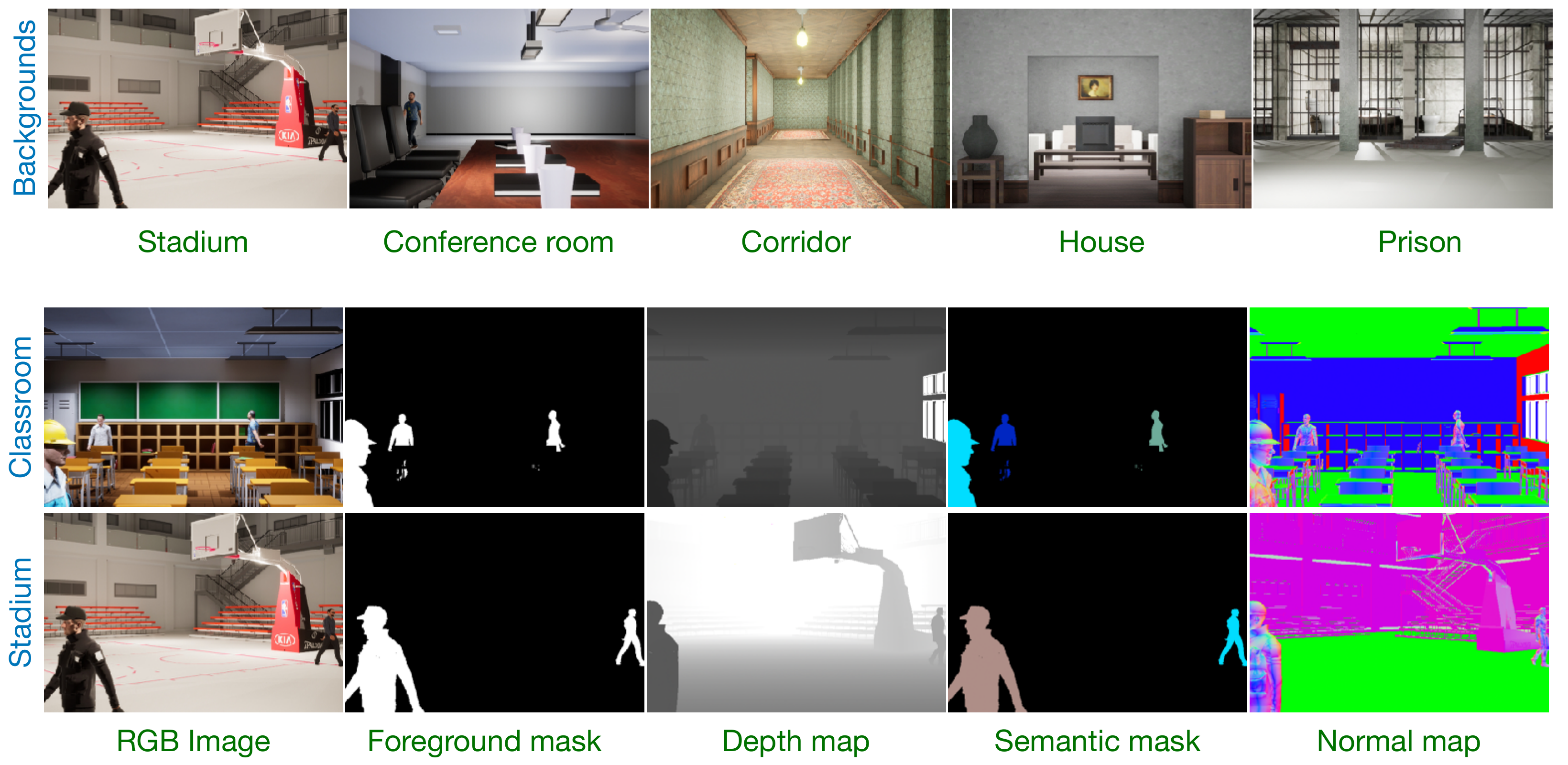}
    \caption{\centering Visualization of sample backgrounds developed in ISD (top row). Visualization of Sample images from {\bf ISD} with their annotations such as binary foreground mask, normal map, depth map, and instance semantic map (bottom two rows). }
    \label{visimgs}
\end{figure*}

\begin{table*}[!t]
\centering
\caption{Dataset description: Indoor Surveillance Data (ISD).}
\resizebox{0.8\textwidth}{!}{
\begin{tabular}{c||c| c| c| c| c | c c} 
\vtop{\hbox{\strut Background}\hbox{\strut Environment}} & $\#$\vtop{\hbox{\strut Lighting}\hbox{\strut Conditions}}  & $\#$ \vtop{\hbox{\strut Camera}\hbox{\strut Angles}} & $\#$ \vtop{\hbox{\strut RGB}\hbox{\strut Images}} & $\#$ \vtop{\hbox{\strut Foreground}\hbox{\strut Masks}}  & $\#$ Depth Maps & $\#$ Normal Maps\\
\hline
\hline
{\it Classroom} & $2$ & $10$  &$20140$ & $20140$ & $20140$ & $20140$\\ 
\hline
{\it Shopping Mall} & $2$ & $8$  &$15984$ & $15984$ & $15984$ & $15984$\\ 
\hline
{\it Prison} & $2$ & $8$ & $16000$  & $16000$ & $16000$ & $16000$\\ 
\hline
{\it Conference Hall} & $2$ & $10$  &$20040$ & $20040$ & $20040$ & $20040$\\ 
\hline
{\it Sport Stadium} & $2$ & $12$  &$24086$ & $24086$ & $24086$ & $24086$\\ 
\hline
{\it Corridor} & $2$ & $8$ & $16112$  & $16112$ & $16112$ & $16112$\\ 
\hline
{\it Metro} & $2$ & $10$ & $20040$  & $20140$ & $20040$ & $20040$\\ 
\hline
{\it House} & $2$ & $9$ & $18036$  & $18036$ & $18036$ & $18036$\\ 
\hline

\end{tabular}}
\label{isd}
\end{table*}

\section{Introduction}
\label{sec:intro}
Video Foreground Segmentation (VFS) \cite{jeeva2015survey,brutzer2011evaluation,benezeth2008review,garcia2020background} aims to segmenting all the moving objects in the video frames, i.e., captured from stationary/nonstationary cameras, and outputs binary foreground
segmentation masks. VFS is an important, but challenging task in many computer vision applications including but not limited to efficient video surveillance\cite{sen2004robust}, motion estimation and anomaly detection\cite{basharat2008learning}, augmented reality\cite{chun2013real}, human tracking\cite{horprasert2000robust}, traffic monitoring\cite{garcia2020background},  and action recognition\cite{zhu2015human}, etc. 

The most common deep learning technique\cite{oktay2018attention,lim2018foreground,lim2020learning} for solving VFS strategy rely on a frame or image as input in a video and does not utilize temporal/motion cues, which is crucial cue for motion-based foreground segmentation. Thus, image-based foreground segmentation techniques, which relies upon appearance features, lack the capability to handle complex scenarios in a general setting. For example, consider the traffic monitoring and video surveillance scenario wherein a robust VFS system needs to perform well on both training data distribution and out-of-distribution data under various challenges, which includes illumination variation, occlusions, dynamic backgrounds like waving tree, rain, snow, air turbulence, camouflage regions, i.e., similarity between foreground pixels and background pixels, camera motions that include camera jittering, camera panning-tilting-zooming. 

The structure of this paper is organized as follows: 
A related literature is briefly reviewed in Section~\ref{relatedwork}. We present the motivation to our work and contribution summary in Section~\ref{motcont}. In Section ~\ref{ISD}, we discuss the steps involved in the data generation pipeline along with the description of the proposed dataset. In Section~\ref{mustanarchs},  we discuss the proposed deep learning architectures. Experimental results corresponding to proposed methods along with benchmarks presented in Section~\ref{expers}. Finally, conclusions are presented in Section~\ref{conc}.

\section{Related Work}
\label{relatedwork}
A large body of literature, which can be categorized into classical approaches and data-driven approaches, is available on {\it motion-based foreground segmentation}. A comprehensive survey of classical computer approaches for motion-based foreground segmentation can be found in \cite{radke2005image,brutzer2011evaluation,benezeth2008review,garcia2020background}. In this section,  we briefly review the most representative deep learning works related to moving object segmentation. 

Classical techniques \cite{radke2005image,brutzer2011evaluation,benezeth2008review,garcia2020background,bouwmans2014traditional,barnich2010vibe,elgammal2002background,stauffer1999adaptive,kim2004background,st2014subsense} on motion-based foreground segmentation have explored the idea of building a background model for a
specific video sequence to identify the foreground objects, i.e., moving objects, using thresholding methods. In past two decades, the parameterized probabilistic models, which operates at a pixel level, have been explored for background subtraction. However, these techniques are computationally expensive. Therefore, recent works on background subtraction are inspired by the nonparametric techniques to overcome the limitation of parametric methods. 

Deep learning architectures have been widely explored for foreground segmentation due to their ability to learn high-level representations from
data\cite{bouwmans2019deep,zeng2018background,chen2017pixelwise,an2023zbs}. Recent deep learning works \cite{sakkos2018end,lim2018foreground,lim2020learning,rahmon2021motion,ronneberger2015u,radke2005image,hou2023survey,icsik2018swcd} have shown to be superior to traditional methods by a significant margin in computer vision tasks. Prior works on foreground segmentation explored  deep learning network architectures such as convolutional neural networks (CNNs)\cite{braham2016deep, zeng2019combining,babaee2018deep,guo2018learning}, generative adversarial networks (GANs) \cite{bakkay2018bscgan}, etc, in supervised learning framework. \cite{braham2016deep} reported the first work that explored deep learning for foreground segmentation.  A patch-based technique, namely DeepBS \cite{babaee2018deep} explored a CNN based architecture and training data includes both input frames and associated background images.  However, these methods have drawbacks such as computationally intensive, overfitting owing to pixel redundancy, loss of contextual information, and requires a lot of patches to train patch-based models\cite{rahmon2021motion}. To alleviate the drawbacks associated with the patch-based methods, \cite{lim2018foreground,lim2020learning,rahmon2021motion} proposed to feed the whole resolution images to the network as input to predict foreground masks. Cascade CNN \cite{wang2017interactive} proposed a cascade structure to synthesize basic CNN and multi-resolution CNNs and trained with specially chosen frames with their ground truth foreground masks. 
\cite{chen2017pixelwise, sakkos2018end} leveraged spatio-temporal cues and these methods followed the training strategy, which involves combining image frames from the video with the generated background models.  BSUV-net \cite{tezcan2019fully} uses a CNN based architecture for background subtraction of unseen videos. BSUV-net input includes  two background frames taken at different time points along with their semantic segmentation masks and current frame. BSGAN\cite{zheng2020novel} exploited the median filtering strategy to estimate the background and utilized in training the Bayesian GAN. 
\cite{ronneberger2015u} proposed UNet architecture for image segmentation and showed promising results for foreground segmentation, which is a special case of image segmentation. \cite{oktay2018attention} introduced attention mechanism to further improve the performance of UNet. Another state-of-the-art network in foreground segmentation is FgSegNet\cite{lim2018foreground,lim2020learning} and it belongs to the category of encoder-decoder networks. FgSegNet and their variants uses three different spatial resolutions of image fed as input to encoder. Further, they use transposed CNNs in the decoder side.\cite{rahmon2021motion} developed deep architectures, namely MU-Net$1$ and MU-Net$2$ for segmenting moving objects. In particular, MU-Net$2$ considered additional cues such as background subtraction mask and flux mask along with current frame as input to improve the foreground accuracy.

\section{Motivation and Contribution}
\label{motcont}
In this section, we present the related works along with their limitations, which motivated this work. Further, contribution is outlined in brief.
\subsection{Motivation}
Existing data-driven techniques \cite{lim2018foreground, lim2020learning, rahmon2021motion, ronneberger2015u,oktay2018attention,zheng2020novel} for motion-based VFS are image-driven and they rely on appearance/spatial cues such as color and texture rather than the relevant attributes such as motion and temporal cues. In particular, a motion-based video foreground segmentation framework that works based on appearance cues suffer from poor performance on out-of-distribution~(OOD) data since appearance cues in test data might differ from training data. FgSegNet and their variants \cite{lim2018foreground, lim2020learning}
are considered as the state-of-the-art for the problem at hand. However, the these models are very big and their out-of-domain data performance is very poor \cite{rahmon2021motion}. To address, \cite{rahmon2021motion} proposed an architecture, namely MU-Net$2$, but it requires extra annotations such as background subtraction mask and flux mask along with RGB image as input. Therefore, getting those annotations during inference are challenging and expensive.

A robust motion-based foreground segmentation needs temporal understanding of video data to capture higher level cues like motion along with the appearance cues. An efficient modelling of motion and temporal cues in VFS helps in overcoming  poor generalizability of image-based methods on OOD video data. In this work, we propose deep models that utilizes temporal cues to improve the robustness of VFS without any additional annotations. There are several benchmark datasets for VFS such as CDnet$2014$ \cite{wang2014cdnet}, SBI$2015$\cite{maddalena2015towards}, BMC$2012$ \cite{vacavant2013benchmark}, and  UCSC \cite{mahadevan2009spatiotemporal}. Although these datasets includes a lot of background environments, but still does not include all possible realistic and challenging backgrounds and foregrounds.~Therefore, having an annotated benchmark dataset that includes complex background environments with complex foregrounds such as number moving objects, camera angles, etc, helps in improving the robustness of data-driven VFS. Therefore, we introduce a novel complex video dataset, which adds diversity and complexity to the existing datasets.

\subsection{Our Contribution}
This work proposes two novel deep learning models that utilizes temporal information and also introduces a new video dataset with multiple frame-level annotations to improve the VFS performance on OOD distribution. In summary, the key contributions are given below: 
\begin{itemize}
    \item[C1.] We propose deep networks, namely {\it MUSTAN1} and {\it MUSTAN2} that utilizes temporal information at various scales in a video data as an attention to obtain an accurate estimation of foreground binary mask. 
    
    \item[C2.] We propose a diverse and complex synthetic video dataset, namely {\it Indoor Surveillance Dataset} (ISD) that includes complex background environments with various camera angles.  ISD has multiple annotations such as foreground binary mask, instance semantic maps, normal maps, and depth maps. Therefore, ISD can be utilized in other computer vision tasks.
    
    \item[C3.] Our methods are validated across various benchmark datasets to establish the fact that proposed temporal modelling indeed contributes to the generizability i.e., superior performance on OOD data, compared to the state-of-the-art.
    
    \item[C4.] Finally, ISD combined with benchmark datasets in training further improves the robustness of VFS.
\end{itemize}

\section{Indoor Surveillance Dataset (ISD)}
\label{ISD}
 To develop an accurate and robust deep learning model for a computer vision task, training data must include complex and diverse scenarios. However, having a real video dataset with the spatio-temporal annotations is a tedious and expensive task. Therefore, a viable solution is to generate  a realistic and high quality synthetic video dataset, which includes complex and diverse environments, with annotations.

\subsection{Data Generation Pipeline}

    A synthetic and realistic video data generation pipeline shown in Fig.~\ref{dgp_} and it is built using Unreal Engine (UE5)\footnote{\label{ue5}https://www.unrealengine.com/en-US/unreal-engine-5} for the use case under consideration. Unreal Engine (UE5)\footref{ue5} is a popular platform in applications such as a virtual or augmented reality, video games, motion capture devices, etc. In UE$5$, we utilize {\it Easysynth}\footnote{https://github.com/ydrive/EasySynth} plugin for generating synthetic images and render the image sequence using game engine rasterizer.  
 
 Generation of a realistic synthetic dataset essentially begins with creating the complex and realistic 3D environments and It includes $3$ major steps: 1. {\bf Environment Creation:} Create 3D environments in UE$5$ by fetching 3D assets from platforms such as {\it sketchfab}\footnote{https://sketchfab.com/} and {\it unreal asset store}\footnote{https://www.unrealengine.com/marketplace/en-US/store} to create complex and realistic environments. These environments serve as background for the generated images. 2. {\bf Scene Composition:} Place 3D humans and objects in a 3D scene. Once the assets are in a place, we position the cameras at various places in the scene to render unique viewpoints along with the specific lighting condition. and 3. {\bf Rendering:} Render the environment using {\it EasySynth}\footnote{https://github.com/ydrive/EasySynth} plugin to generate RGB images with frame-level annotations such as binary foreground mask, depth, normal, and instance segmentation maps. The compute that we utilized for the above framework is {\it Windows $11$} PC with {\it Nvidia RTX$3050$}.

\subsection{Dataset Description}

 There are state-of-the-art benchmarking datasets such as CDnet$2014$\cite{wang2014cdnet} and SBI$2015$ \cite{maddalena2015towards} for foreground segmentation. However, these datasets lack challenging backgrounds and foregrounds with various camera viewpoints for indoor surveillance. To address above limitations, we propose a new synthetic video dataset, namely {\it Indoor Surveillance Dataset} (ISD) for motion-based video foreground segmentation to bring diversity and complexity on top of the existing benchmark datasets. A brief summary of ISD is given below:
 \begin{itemize}
     \item The ISD is composed of nearly $1,50,538$ high-quality RGB images that are of realistic looking and having multiple annotations for moving objects such as foreground binary mask, instance segmentation mask, normal maps, and depth maps ( Fig.~\ref{visimgs}).
     \item ISD has $8$ backgrounds along with $8$ to $12$ camera views for each background. Further, it has two lighting conditions such as daylight and night light. Summery of the ISD dataset is provided in Table~\ref{isd} and few sample images from ISD are illustrated in Fig.~\ref{visimgs} for visualization.
 \end{itemize}
 
  ISD also benefits computer vision tasks such as depth map prediction, and instance segmentation along with VFS.

\section{Proposed Methodology}
\label{mustanarchs}
We propose novel deep learning architectures for robust
motion-based video foreground segmentation. Our architectures make use of multi-scale temporal information along with the spatial cues
to boost the foreground segmentation accuracy on OOD. Our architectures, namely {\it MUSTAN1} and {\it MUSTAN2} belong to the class of encoder-decoder type of architecture, but trained with two variations in the input streams along with custom attention derived based on temporal embeddings.
\subsection{MUSTAN1}
\begin{figure*}[!t]
    \centering
    \includegraphics[scale=0.18]{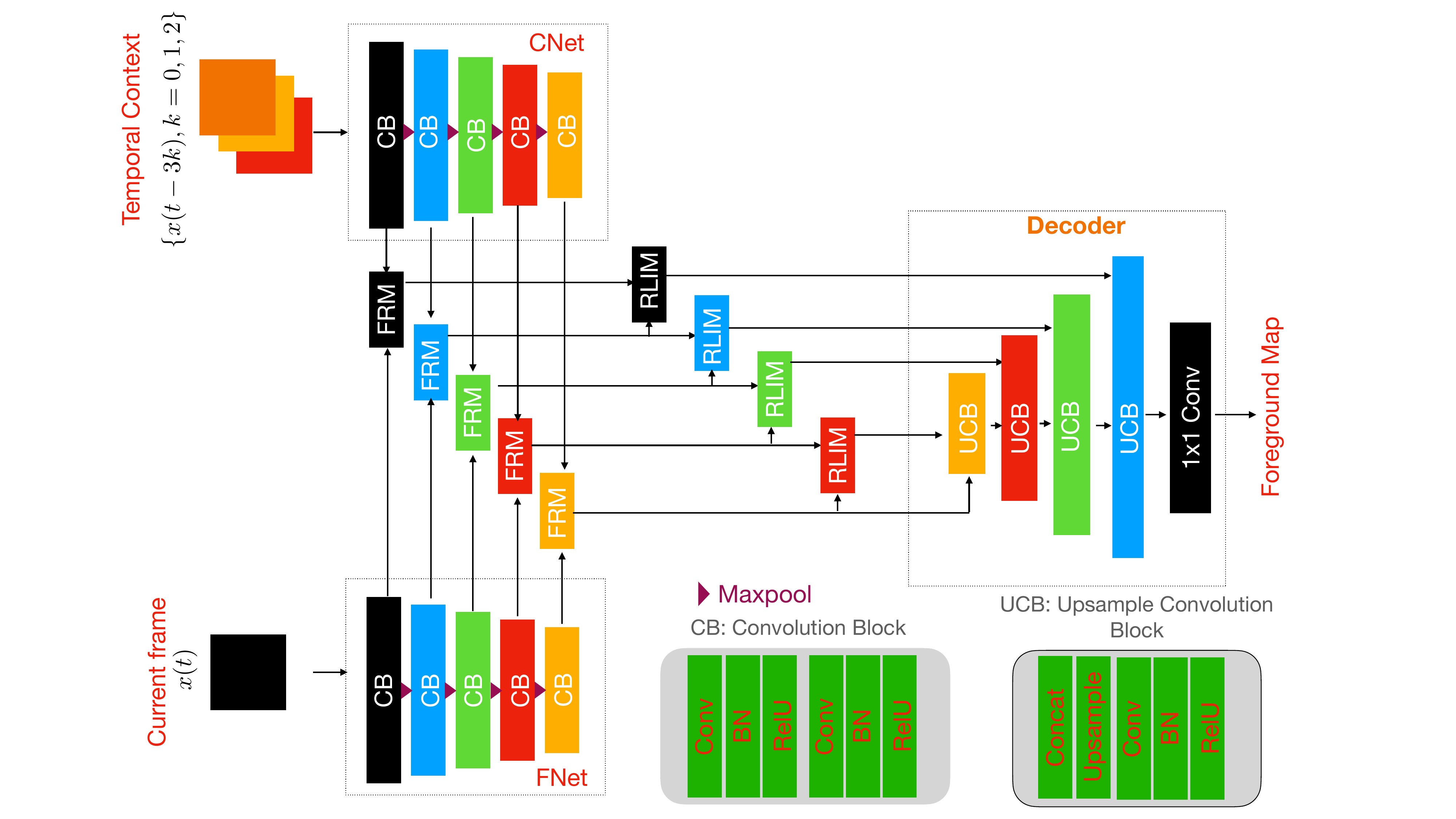}
    \caption{\centering {\it MUSTAN1} architecture. BN stands for batchnorm layer, "Conv" denotes convolutional layer, CNet denotes Context Network, FNet stands for Feature Network, FRM stands for Feature Refinement Module, and RLIM denotes Refine Localization Information Module. }
    \label{mastan1}
\end{figure*}

MUSTAN1 (Fig.~\ref{mastan1}) consisting of two encoders and one decoder wherein one encoder is called context network (CNet), i.e., extracts temporal embeddings at multiple scales, and the other one is feature network (FNet), which extracts the spatial cues from current frame. Further, attention is derived from multi-scale temporal embeddings to highlight salient portions of current frame representations at different scales before they fed to decoder.

\begin{figure}[!t]
    \centering
    \includegraphics[width = 0.95\linewidth]{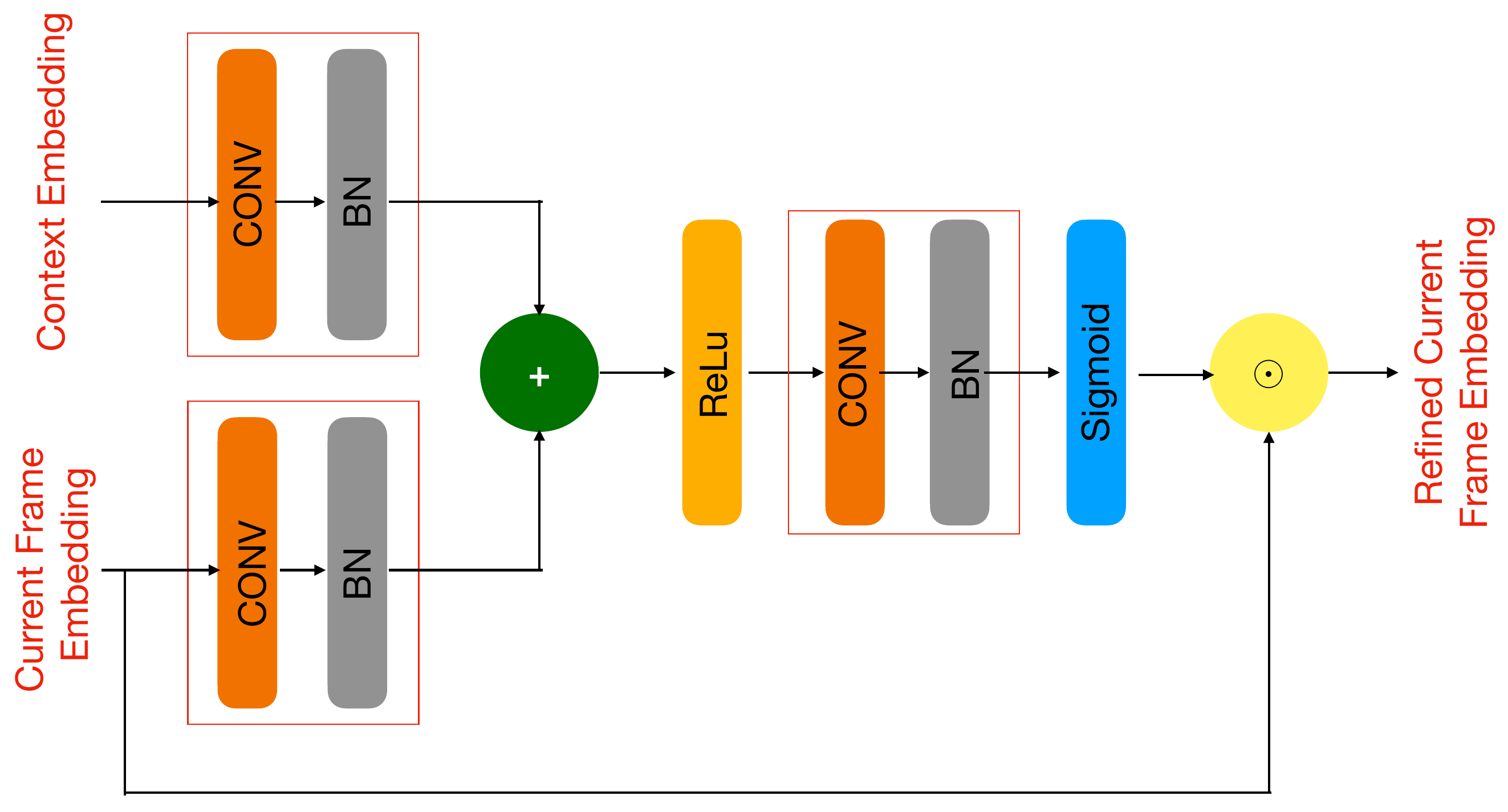}
    \caption{\centering Feature Refinement Module (FRM). {\bf CONV} stands for convolution layer, {\bf BN} stands for batch normalization layer, $+$ denotes element wise addition, and $\cdot$ denotes element wise multiplication. }
    \label{frm}
\end{figure}

\begin{figure}[!t]
    \centering
    \includegraphics[width = 0.95\linewidth]{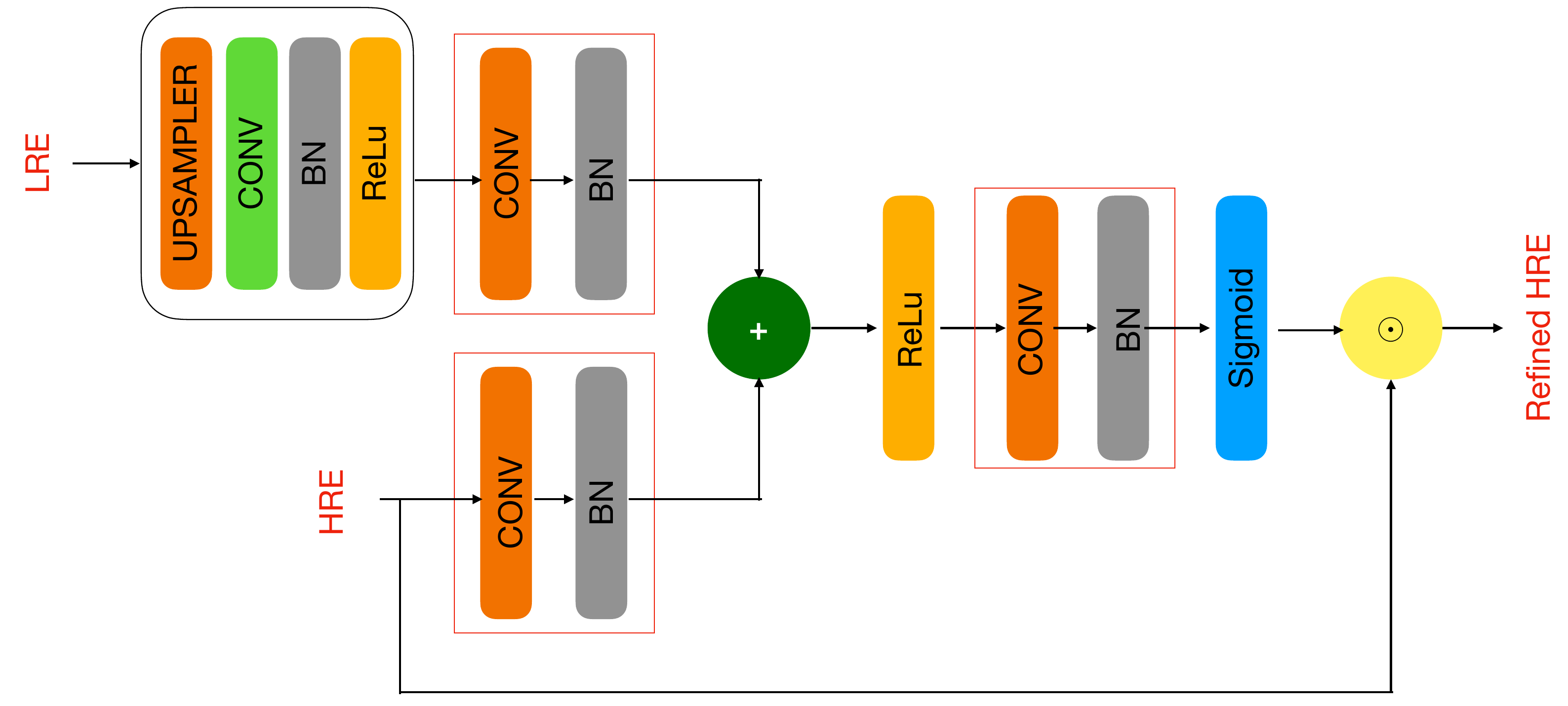}
    \caption{\centering Refine Localization Information Module (RLIM). {\bf LRE} stands for low resolution embedding, {\bf HRE} stands for high resolution embedding.}
    \label{rlim}
\end{figure}

To ensure real-time inference, encoders of our models, namely CNet and FNet are developed based on the five blocks of ResNet$18$ \cite{he2016deep} and number of convolutional kernels in each of these blocks are $64, 128, 256, 512,$ and $1024$, respectively. However, both CNet and FNet differ only in number of input channels, that is, input dimension of CNet is $3*T\times 320 \times 480$, $T$ is number of frames whereas input dimension of FNet is $3\times320\times480$. In the proposed architecture, we introduced two modules, namely Feature Refinement Module (FRM) (Fig.~\ref{frm}) and Refine Localization Information Module (RLIM)(Fig.~\ref{rlim}) that are inspired by the attention mechanism introduced in \cite{oktay2018attention}. FRM takes both temporal context embedding and current frame embedding as input and then highlights the relevant portions of current frame embedding based on attention weights derived as shown in Fig.~\ref{frm}. These refined current frame embedding are utilized in skip connections. Whereas RLIM takes low resolution and high resolution embeddings of current frame and refines the semantic information present in the high resolution feature map.

The decoder of our model has $4$ blocks, where each block receives the feature maps that are upsampled and concatenated with the embedding coming from the  RLIM in the respective skip connection. Then, concatenated embedding is fed to the $3 \times 3$ convolution layer followed by a ReLu activation layer. The output of the final feature extraction block of decoder is fed to $1\times1$ convolution layer. which decreases the number of feature maps, followed by a sigmoid activation layer to produces the class label probabilities.

\subsection{MUSTAN2}

\begin{figure*}[!t]
    \centering
    \includegraphics[scale=0.18]{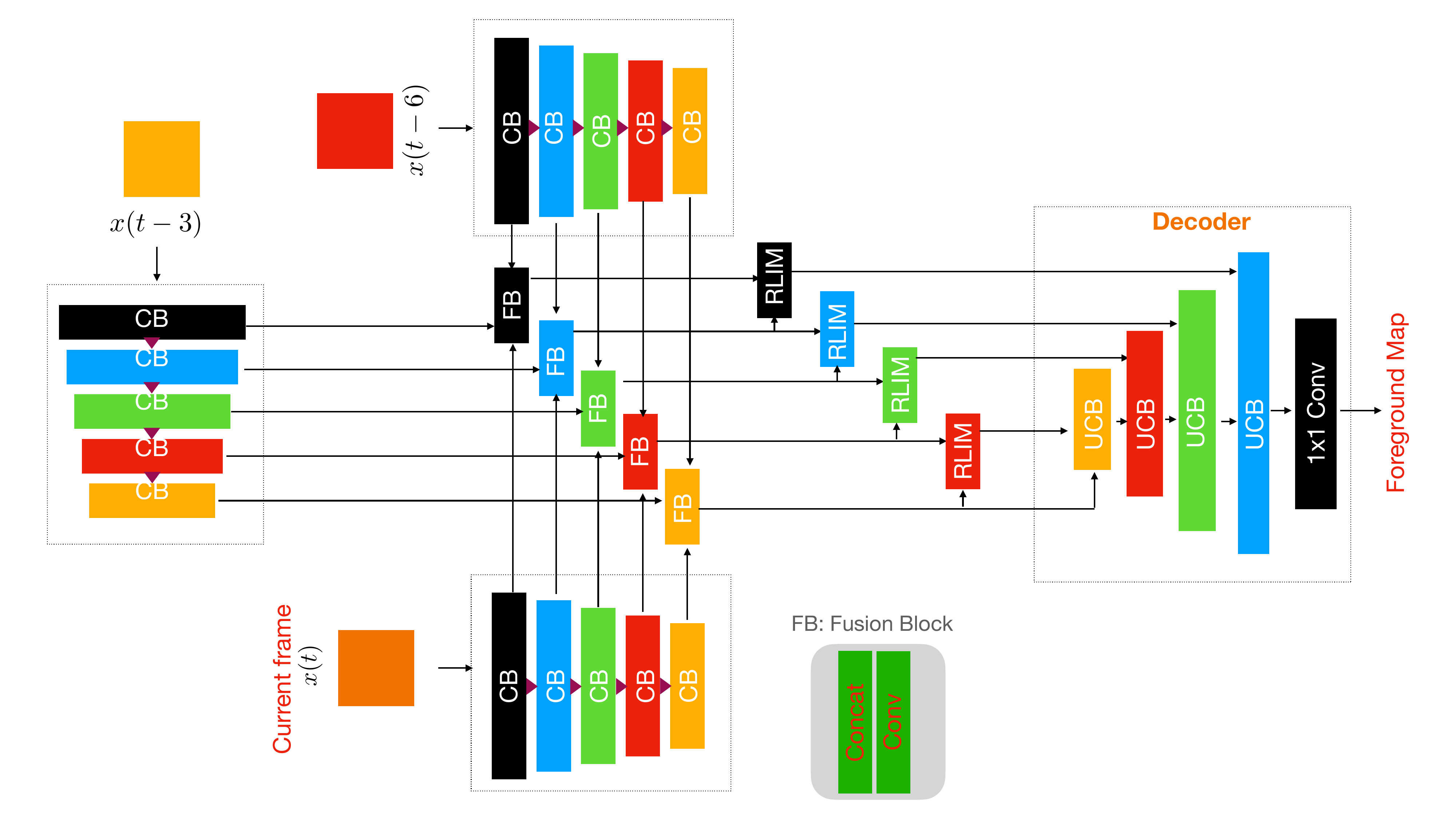}
    \caption{\centering {\it MUSTAN2} architecture. FB stands for Fusion Module, and RLIM denotes Refine Localization Information Module. }
    \label{mastan2}
\end{figure*}

We used temporal embedding to refine current feature embedding through FRM in MUSTAN1. MUSTAN1 is one of the possible ways of modelling the temporal information. Therefore, we explore another temporal modelling strategy based on mid-level fusion that lead to a novel deep learning architecture, namely MUSTAN2.
Fig.~\ref{mastan2} illustrates the architecture of MUSTAN$2$
with a multi-scale fusion of temporal information and feature enhacement through RLIM module at various scales. MUSTAN$2$ consisting of three encoders and one decoder. Each encoder of MUSTAN$2$ takes one frame within the temporal window as input and extracts correpsonding feature representation at multiple scales. Then, feature maps of frames within the temporal window is fed to fusion block (FB) (Fig.~\ref{mastan2}) in the skip connection to decrease the
number of feature maps and to have a common embedding that captures temporal information. Further, RLIM takes FB output and low resolution embedding of current frame as input and outputs refined high resolution feature map that is fed to decoder. The decoder of MUSTAN$2$ is similar to that of MUSTAN$1$ and it outputs the class label probabilities as output.


\section{Experimental Results and Analysis}
\label{expers}
In this section, we describe the datasets, present the experimental settings along with evaluation metrics. Further, we present the qualitative and quantitative analysis of our deep learning architectures, namely MUSTAN$1$ and MUSTAN$2$ compared to the state-of-the-art.

\begin{figure}[!t]
    \centering
    \includegraphics[scale=0.21]{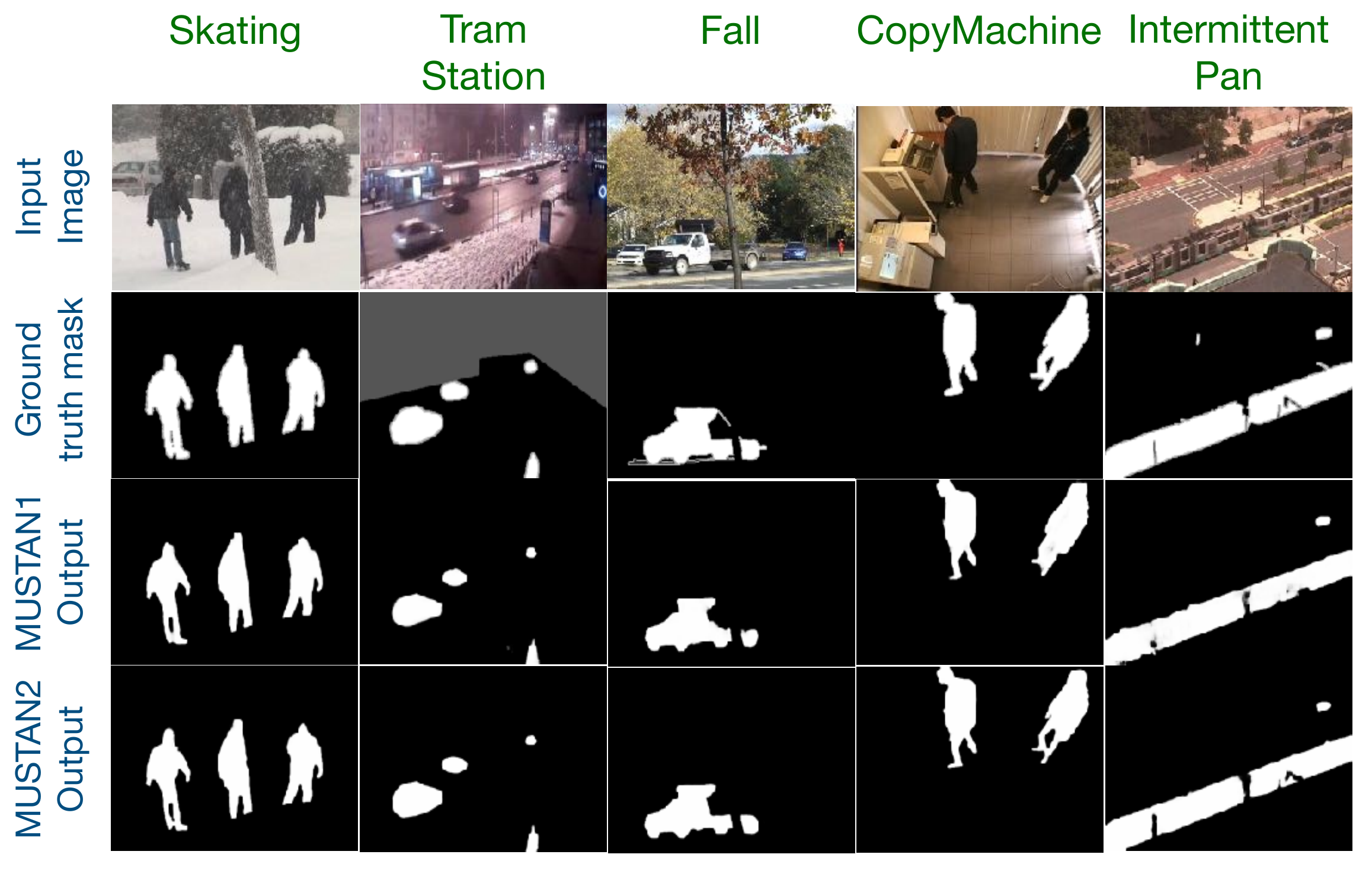}
    \caption{\centering Visual illustration of In-Domain performance of MUSTAN1 and MUSTAN2: Proposed models trained on subset of CD$2014$ and tested on subset of CD$2014$. Each column is a particular video category available in CD$2014$.}
    \label{cdvis}
\end{figure}
\begin{figure}[!t]
    \centering
    \includegraphics[scale=0.21]{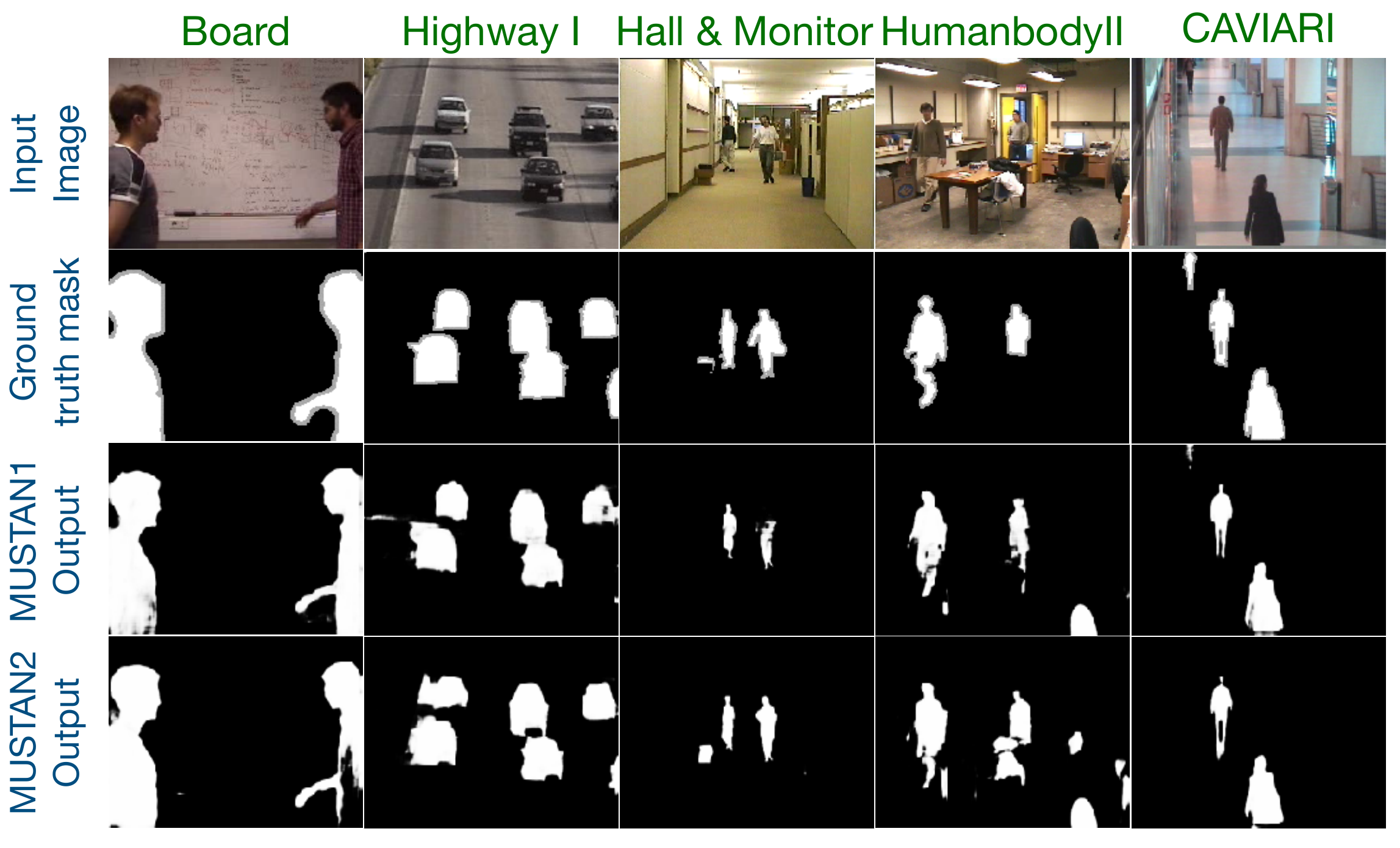}
    \caption{\centering Visual illustration of OOD performance of MUSTAN1 and MUSTAN2: Proposed models trained on CD$2014$ and tested on SBI$2015$. Each column is a particular video category available in SBI$2015$.}
    \label{sbivis}
\end{figure}

\subsubsection{Baselines}

We compare our methods with the various state-of-the-art supervised
deep learning methods such as UNet \cite{ronneberger2015u}, Attention UNet \cite{oktay2018attention}, Cascade CNN \cite{wang2017interactive}, BSPVGAN \cite{zheng2020novel}, BSGAN \cite{zheng2020novel}, FgSegNet \cite{lim2018foreground}, FgSegNet(S) \cite{lim2020learning}, FgSegNet(v2) \cite{lim2020learning}, MU-Net$1$ \cite{rahmon2021motion}, and MU-Net$2$ \cite{rahmon2021motion}. The source codes for FgSegNet, MU-Net1, and MU-Net2 available online \footnote{\label{ds}https://github.com/lim-anggun/}\textsuperscript{,}\footnote{https://github.com/CIVA-Lab/Motion-U-Net}. Our experimental settings are same as recommended in \cite{lim2020learning,rahmon2021motion} for fair evaluation.

\subsubsection{Datasets}
We consider the proposed dataset, namely ISD along with state-of-the-art datasets such as scene background initialization (SBI$2015$) dataset \cite{maddalena2015towards}, which contains $14$ labelled video sequences, and change detection challenge (CDnet$2014$) dataset\cite{wang2014cdnet} to asses the out-of-domain or generalization performance of proposed models. CDnet$2014$ has $11$ realistic and challenging categories, i.e., illumination change, shadow, dynamic background motion and camera motion, etc.,  wherein each category contains $4$ to $6$ video sequences. Therefore, it has a total of $53$ video sequences, which include $160$K frames and $118$K labeled frames with the spatial resolutions vary from $320\times240$ to $720\times576$.

\subsection{Evaluation Metrics}
We compute the metrics such as Precision (Pr), Recall (Re), specificity (Sp), and F1 score to evaluate both In-Domain and Out-Of-Domain performance of the proposed and benchmark methods.

\begin{eqnarray}
   \text{Pr} = \frac{TP}{TP+FP}, ~ ~ ~ \, \, \,  \text{Re} = \frac{TP}{TP+TN},\\
     \text{Sp} = \frac{TP}{TP+FP}, ~ ~ ~ \, \, \,  \text{F1 Score} = \frac{2*Pr*Re}{Pr+Re},
\end{eqnarray}

where TN stands for true negatives, FP denotes false positives, FN refers to false negatives, and TP stands for true positives.

\subsection{Hyper-parameter Settings}

Encoders of proposed networks are based on ResNet$18$\cite{he2016deep} and initalized with ImageNet pretrained weights. Spatial resolution of input RGB frames is set to $320 \times 480$. The optimizer used is Adam with learning rate initialized to $1e^{-4}$. The learning scheduler, namely {\it StepLR} is considered and the parameters such as step size and gamma are set to $20$ and $0.1$, respectively. Thus, learning scheduler reduces the learning rate by a gamma factor for every $20$ epochs, which is determined by the step size. The dataset is shuffled and then split into the ratio of $90\%:10\%$ for training and validation, respectively. Our models are trained for $40$ epochs with a batch size equals to $8$. Further, the loss function considered in training our networks is defined below:
\begin{equation}
    L(\boldsymbol{y}, \boldsymbol{\hat{y}}, \boldsymbol{p}) = \theta * L_{\text{tl}} + (1-\theta) * L_{\text{bce}},
\end{equation}

where $*$ denotes multiplication operation, $L_{\text{tl}}$ denotes Tversky loss (TL) \cite{salehi2017tversky}, $L_{\text{bce}}$ refers to the binary cross-entropy (BCE) loss and $\theta$, which is set to $0.5$ in our experiments, is weight parameter, i.e., determines trade-off between $L_{\text{tl}}$ and $L_{\text{bce}}$. 
TL (Eq.~\ref{tl}) results in a better trade-off between precision and recall.
\begin{equation}
\label{tl}
    L_{\text{tl}}(\boldsymbol{y}, \boldsymbol{\hat{y}}, \alpha, \beta) = \frac{TP}{TP + \alpha FP + \beta FN},
\end{equation}
where $\boldsymbol{p}$ network output probabilities, $\boldsymbol{y}$ is the binary ground-truth foreground mask, $\boldsymbol{\hat{y}} \in \{0, 1\} $ is the predicted binary foreground mask, which is obtained after thresholding $\boldsymbol{p}$, $\alpha, \beta$ are weights associated with FPs and FNs, TP = $|\boldsymbol{y}\cdot \boldsymbol{\hat{y}}|$, FP = $|(1-\boldsymbol{y})\cdot \boldsymbol{\hat{y}}|$, FN=$|\boldsymbol{y}\cdot (1-\boldsymbol{\hat{y}})|$, $|\cdot|$ stands for cardinality measure, and $\cdot$ denotes element-wise multiplication. BCE loss $L_{\text{bce}}$ is defined as
\begin{equation}
\label{bce}
    L_{\text{bce}}(\boldsymbol{y}, \boldsymbol{p}) = -(\boldsymbol{y}*\log(\boldsymbol{p}) + (1-\boldsymbol{y})*\log(1-\boldsymbol{p})).
\end{equation}

\subsection{Quantitative and Qualitative Results}
We trained proposed networks, namely MUSTAN1 and MUSTAN2 on the
challenging subset of CD$2014$ dataset. The training/testing splits
in our experiments are same as the ones recommended in \cite{lim2020learning,lim2018foreground,rahmon2021motion} and those are available online\footref{ds}. Training data ($\approx10K$), which is a subset of CDnet$2014$, consists of $200$ labeled frames from each video sequence present in CDnet$2014$ dataset.

To asses the In-Domain performance of our methods along with benchmarks, we considered the test set\footref{ds} from the CDnet$2014$. Test data is based on $25$ to $50$ per video sequence from the $11$ categories present in CDnet$2014$. The images considered in the test split are not part of train split. Few sample results of our methods are illustrated in Fig.~\ref{cdvis}. One can observe that our methods produces accurate foreground segmentation masks on OOD data. Further, the performance metric, namely F1 score is computed (on the test set) for each video category of CDnet$2014$ and is reported in Table~\ref{idf1}. We can infer that proposed architecture superior to the state-of-the-art in terms of average F1 score. Table~\ref{cd_avgf1} illustrates the performance of all the methods in terms of evaluation metrics mentioned above. As can be seen, our methods outperformed (in terms of F1 score) the state-of-the-art except FgSegNet. One important observation from Table~\ref{sbi_ood} and Table~\ref{modelsize} is that the FgSegNet is a very big model and significantly underperforms on Out-Of-Domain data since it is overfitting the In-Domain data. 

To asses the Out-Of-Domain (OOD) performance of our methods along with the benchmarks, we trained models on CDnet$2014$ and the tested on $7$ categories videos in SBI$2015$. Detailed OOD performance (considering $7$ videos of SBI$2015$ for testing) of our methods is reported in Table~\ref{ood1}, Table~\ref{ood2} and Table~\ref{ood_isd}. OOD performance, i.e., in terms of Avg. F1 score, of our methods and competing methods is illustrated in Table~\ref{sbi_ood}. As can been seen, proposed methods are superior to benchmarks by some margins on OOD data.  Table~\ref{ood2} and Table~\ref{ood_isd} demonstrate that ISD results in OOD performance gain.

\begin{table}[!t]
\centering
\caption{In-domain performance, i.e., measured in terms of F1-score, of {\it MUSTAN1} (Ours) and {MUSTAN2} (Ours) compared to baselines on CD2014 dataset \cite{wang2014cdnet}. BW: badWeather, BL: baseline, CJ: cameraJit., DB: dynamicBg., IM: intermittent object motion, LFR: lowFrameR., NV: nightVid., SD: shadow, Ther.: thermal, Tur.: turbulence, and Avg. denotes Average.}
\resizebox{0.5\textwidth}{!}{
\begin{tabular}{c|| c| c| c |c |c } 
Video & FTSG & MU-Net1 &  MU-Net2 & MUSTAN1 & MUSTAN2 \\
\hline
\hline

{\it PTZ} & $0.3241$ & $0.7946$ & $0.8185$ & $0.8985$ & $0.9354$ \\ 

{\it BW} & $0.8228$  & $0.9319$ & $0.9343$ & $0.9553$  & $0.9730$\\ 

{\it BL} & $0.9330$  & $0.9875$ & $0.9900$ & $0.9304$  & $0.9537$\\ 

{\it CJ} & $0.7513$  & $0.9802$ & $0.9824$ & $0.9383$  & $0.9572$\\ 

{\it DB}&  $0.8792$ & $0.9836$ & $0.9892$ & $0.9233$  & $0.9551$ \\ 

{\it IM} & $0.7891$  & $0.9872$ & $0.9894$ & $0.9291$  & $0.9646$\\

{\it LFR} & $0.6259$  & $0.7237$ & $0.8706$ & $0.8391$  & $0.9113$\\

{\it NV} & $0.5130$  & $0.8575$ & $0.8362$&  $0.8954$ & $0.9513$\\

{\it SD} &  $0.8535$ & $0.9825$ & $0.9845$&  $0.9420$ & $0.9670$ \\
{\it Ther.} & $0.7768$  & $0.9825$ & $0.9842$& $0.9248$  & $0.9574$\\
{\it Tur.} & $0.7127$  & $0.8499$ & $0.9272$ & $0.9094$  & $0.9395$ \\
\hline
\hline
{\it Avg.} & $0.7283$  & $0.9147$ & $0.9369$ &  $0.9168$ & $\mathbf{0.9514}$\\
\hline
\hline
\end{tabular}}
\label{idf1}
\end{table}

\begin{table}[!t]
\centering
\caption{Overall average Performance, i.e., in terms of F1 score, Precision, and Recall, of {\it MUSTAN1} and {MUSTAN2} compared to baselines across all the video categories of CD2014 dataset\cite{wang2014cdnet}.}
\resizebox{0.4\textwidth}{!}{
\begin{tabular}{c|| c| c| c} 
Method & {\it F$1$} Score & Precision &  Recall\\
\hline
\hline
{\it Cascade CNN}\cite{wang2017interactive} & $0.9209$  & $0.8997$  & $0.9506$ \\ 

{\it BSGAN}\cite{zheng2020novel} & $0.9339$ & $0.9232$ & $0.9476$ \\ 

{\it BSPVGAN}\cite{zheng2020novel} & $0.9472$  & $0.9501$ & $0.9544$ \\ 

{\it FgSegNet}\cite{lim2018foreground} & $0.9770$  & $0.9758$ & $0.9836$ \\ 

{\it FgSegNet(S)}\cite{lim2020learning} & $0.9804$  & $0.9751$ & $0.9896$ \\ 

{\it FgSegNet(v2)}\cite{lim2020learning}&  $0.9847$ & $0.9823$ & $0.9891$ \\ 

{\it MUNet1}\cite{rahmon2021motion} & $0.9147$  & $0.9414$ & $0.9277$\\

{\it MUNet2}\cite{rahmon2021motion} & $0.9369$  & $0.9407$ & $0.9454$\\

{\it MUSTAN1 (Ours)} & $0.9168$  & $0.8659$ & $0.9417$ \\

{\it MUSTAN2 (Ours)} & $0.9514$  & $0.9156$  & $0.9574$ \\
\hline
\end{tabular}}
\label{cd_avgf1}
\end{table}

\begin{table}[!t]
\centering
\caption{Out-Of-Domain (OOD) Performance: {\it MUSTAN1} and {MUSTAN2} architectures trained on CDnet2014 \cite{wang2014cdnet} and tested on SBI2015 dataset \cite{maddalena2015towards}. }
\resizebox{0.95\linewidth}{!}{
\begin{tabular}{c|| c| c| c} 
Method & {\it F$1$} Score & Precision &  Recall\\
\hline
\hline
{\it UNet}\cite{ronneberger2015u} & $0.4344$ & $0.3656$ & $0.7095$\\
{\it Attention UNet} \cite{oktay2018attention} & $0.5048$ & $0.4977$ & $0.7357$\\

{\it MUNet1}\cite{rahmon2021motion} & $0.3785$  & $0.2881$ & $0.8094$ \\ 

{\it MUNet2} \cite{rahmon2021motion} & $0.7625$  & $0.8484$ & $0.7302$\\

{\it FgSegNet(v2) ($50\%$)}\cite{lim2020learning}& $0.3519$  & $0.8150$  & $0.2419$ \\ 

{\it MUSTAN1 (Ours)} & $0.7635$  & $0.7073$ & $0.8296$ \\

{\it MUSTAN2 (Ours)} & $\mathbf{0.7933}$  & $0.7484$ & $\mathbf{0.8613}$\\
\hline
\end{tabular}}
\label{sbi_ood}
\end{table}

\begin{table}[!t]
\centering
\caption{OOD Performance of {\it MUSTAN1} on SBI$2015$ Dataset \cite{maddalena2015towards}.}
\resizebox{0.45\textwidth}{!}{
\begin{tabular}{c|| c| c| c| c} 
Video &  {\it F$1$} Score & Precision & Specificity & Recall\\
\hline
\hline
{\it HighwayI} & $0.8349$ & $0.8495$ & $0.9852$ & $0.8207$\\ 
\hline
{\it Board} &  $0.8898$ & $0.9143$ & $0.9696$ & $0.8666$\\ 
\hline
{\it HallAndMonitor} &  $0.7033$ & $0.8412$ & $0.9975$ & $0.6043$\\ 
\hline
{\it CAVIAR1} &  $0.8707$ & $0.8079$ & $0.9925$ & $0.9439$\\ 
\hline
{\it HumanBody2} &  $0.6928$ & $0.6565$ & $0.9603$ & $0.7333$\\ 
\hline
{\it HighwayII}&  $0.6124$ & $0.4651$ & $0.9721$ & $0.8962$\\ 
\hline
{\it IBMtest2} &  $0.5777$ & $0.4166$ & $0.9385$ & $0.9423$\\
\hline
\end{tabular}}
\label{ood1}
\end{table}


\begin{table}[!t]
\centering
\caption{OOD Performance of {\it MUSTAN2} on SBI$2015$ Dataset \cite{maddalena2015towards}. {\it MUSTAN2} trained on CDnet$2014$}
\resizebox{0.45\textwidth}{!}{
\begin{tabular}{c||c| c| c| c} 
Video &  {\it F$1$} Score & Precision & Specificity & Recall\\
\hline
\hline
{\it HighwayI} &  $0.8239$ & $0.8274$ & $0.9826$ & $0.8204$\\ 
\hline
{\it Board} &  $0.8729$ & $0.9124$ & $0.9700$ & $0.8367$\\ 
\hline
{\it HallAndMonitor} &  $0.7926$ & $0.7360$ & $0.9932$ & $0.8586$\\ 
\hline
{\it CAVIAR1} &  $0.8682$ & $0.8432$ & $0.9945$ & $0.8947$\\ 
\hline
{\it HumanBody2} &  $0.6969$ & $0.5746$ & $0.9322$ & $0.8854$\\ 
\hline
{\it HighwayII}&  $0.7613$ & $0.7452$ & $0.9928$ & $0.7782$\\ 
\hline
{\it IBMtest2} &  $0.7373$ & $0.6004$ & $0.9704$ & $0.9551$\\
\hline
\end{tabular}}
\label{ood2}
\end{table}

\begin{table}[!t]
\centering
\caption{OOD Performance of {\it MUSTAN2} on SBI$2015$ Dataset \cite{maddalena2015towards}. {\it MUSTAN2} trained on CDnet$2014$ + ISD}
\resizebox{0.45\textwidth}{!}{
\begin{tabular}{c||c| c| c| c} 
Video &  {\it F$1$} Score & Precision & Specificity & Recall\\
\hline
\hline
{\it HighwayI} &  $0.8097$ & $0.8299$ & $0.9835$ & $0.7904$\\ 
\hline
{\it Board} &  $0.9152$ & $0.8878$ & $0.9554$ & $0.9443$\\ 
\hline
{\it HallAndMonitor} &  $0.7856$ & $0.7927$ & $0.9955$ & $0.7787$\\ 
\hline
{\it CAVIAR1} &  $0.8374$ & $0.8470$ & $0.9950$ & $0.8280$\\ 
\hline
{\it HumanBody2} &  $0.8419$ & $0.8018$ & $0.9773$ & $0.8863$\\ 
\hline
{\it HighwayII}&  $0.8350$ & $0.7597$ & $0.9920$ & $0.9268$\\ 
\hline
{\it IBMtest2} &  $0.8872$ & $0.8198$ & $0.9901$ & $0.9667$\\
\hline
\end{tabular}}
\label{ood_isd}
\end{table}

\begin{table}[!t]
\centering
\caption{Model complexity: Network size in terms of number of parameters. M stands for million.}
\resizebox{0.75\linewidth}{!}{
\begin{tabular}{c|| c} 
Model & Number of Parameters  \\
\hline
\hline
{\it MUNet1} & $17.8$M   \\ 

{\it MUNet2} & $17.8$M  \\

{\it FgSegNet(v2)}& $489$M   \\ 

{\it MUSTAN1 (Ours)} & $55.13$M   \\

{\it MUSTAN2 (Ours)} & $40.85$M  \\
\hline
\end{tabular}}
\label{modelsize}
\end{table}

\section{Conclusions}
\label{conc}
We proposed deep learning architecture, i.e., MUSTAN1 and MUSTAN2, that captures temporal cues of video data for enhanced OOD performance in VFS. We also introduced a new dataset, namely Indoor Surveilance Data (ISD) that brings diversity and complexity to the benchmark datasets.  Experiments on benchmark datasets demonstrate that our methods are superior to competing techniques. To put the work in context, we reported precision, recall, specificity, and F1 scores to indicate that our approaches efficient compared to the baselines. Our models are compact and superior to FgSegNet on OOD data. Further, MUSTAN$2$ is superior to MU-Net$2$ and also it does not need extra annotations as input.  Future work involves exploring other cues like depth along with temporal cues to develop a robust VFS model based on MUSTAN1 and MUSTAN2.

{\small
\bibliographystyle{ieee_fullname}
\bibliography{egbib}

\begin{thebibliography}{10}\itemsep=-1pt

\bibitem{an2023zbs}
Y. An, X. Zhao, T. Yu, H. Guo, C. Zhao, M. Tang, and J. Wang.
\newblock {ZBS}: {Z}ero-shot background subtraction via instance-level
  background modeling and foreground selection.
\newblock {\em Proceedings of the IEEE/CVF Conference on Computer Vision and
  Pattern Recognition}, pages 6355--6364, 2023.

\bibitem{babaee2018deep}
M. Babaee, D.~T. Dinh, and G. Rigoll.
\newblock A deep convolutional neural network for video sequence background
  subtraction.
\newblock {\em Pattern Recognition}, 76:635--649, 2018.

\bibitem{bakkay2018bscgan}
M.~C. Bakkay, H.~A. Rashwan, H. Salmane, L. Khoudour, D. Puig, and Y. Ruichek.
\newblock {BSCGAN}: {D}eep background subtraction with conditional generative
  adversarial networks.
\newblock {\em in the IEEE Proceedings of International Conference on Image
  Processing}, pages 4018--4022, 2018.

\bibitem{barnich2010vibe}
O. Barnich and M. Van~Droogenbroeck.
\newblock Vi{B}e: A universal background subtraction algorithm for video
  sequences.
\newblock {\em IEEE Transactions on Image Processing}, 20(6):1709--1724, 2010.

\bibitem{basharat2008learning}
A. Basharat, A. Gritai, and M. Shah.
\newblock Learning object motion patterns for anomaly detection and improved
  object detection.
\newblock {\em Proceeding of IEEE International Conference on Computer Vision
  and Pattern Recognition}, pages 1--8, 2008.

\bibitem{benezeth2008review}
Y. Benezeth, P.-M. Jodoin, B. Emile, H. Laurent, and C. Rosenberger.
\newblock Review and evaluation of commonly-implemented background subtraction
  algorithms.
\newblock {\em 19th International Conference on Pattern Recognition}, pages
  1--4, 2008.

\bibitem{bouwmans2014traditional}
T. Bouwmans.
\newblock Traditional and recent approaches in background modeling for
  foreground detection: {A}n overview.
\newblock {\em Computer Science Review}, 11:31--66, 2014.

\bibitem{bouwmans2019deep}
T. Bouwmans, S. Javed, M. Sultana, and S.~K. Jung.
\newblock Deep neural network concepts for background subtraction: A systematic
  review and comparative evaluation.
\newblock {\em Neural Networks}, 117:8--66, 2019.

\bibitem{braham2016deep}
M. Braham and Marc Van~D.
\newblock Deep background subtraction with scene-specific convolutional neural
  networks.
\newblock {\em International Conference on Systems, Signals and Image
  processing}, pages 1--4, 2016.

\bibitem{brutzer2011evaluation}
S. Brutzer, B. H{\"o}ferlin, and G. Heidemann.
\newblock Evaluation of background subtraction techniques for video
  surveillance.
\newblock {\em In Proceedings of International Conference on Computer Vision
  and Pattern Recognition}, pages 1937--1944, 2011.

\bibitem{chen2017pixelwise}
Y. Chen, J. Wang, B. Zhu, M. Tang, and H. Lu.
\newblock Pixelwise deep sequence learning for moving object detection.
\newblock {\em IEEE Transactions on Circuits and Systems for Video Technology},
  29(9):2567--2579, 2017.

\bibitem{chun2013real}
W.~H. Chun and T. H{\"o}llerer.
\newblock Real-time hand interaction for augmented reality on mobile phones.
\newblock pages 307--314, 2013.

\bibitem{elgammal2002background}
A. Elgammal, R. Duraiswami, D. Harwood, and L.~S. Davis.
\newblock Background and foreground modeling using nonparametric kernel density
  estimation for visual surveillance.
\newblock {\em Proceedings of the IEEE}, 90(7):1151--1163, 2002.

\bibitem{garcia2020background}
B. Garcia-Garcia, T. Bouwmans, and A.~J.~R. Silva.
\newblock Background subtraction in real applications: Challenges, current
  models and future directions.
\newblock {\em Computer Science Review}, 35:100204, 2020.

\bibitem{guo2018learning}
E. Guo, X. Fu, J. Zhu, M. Deng, Y. Liu, Q. Zhu, and H. Li.
\newblock Learning to measure change: Fully convolutional siamese metric
  networks for scene change detection.
\newblock {\em arXiv:1810.09111}, 2018.

\bibitem{he2016deep}
K. He, X. Zhang, S. Ren, and J. Sun.
\newblock Deep residual learning for image recognition.
\newblock {\em Proceedings of the IEEE Conference on Computer Vision and
  Pattern Recognition}, pages 770--778, 2016.

\bibitem{horprasert2000robust}
T. Horprasert, D. Harwood, and L.~S. Davis.
\newblock A robust background subtraction and shadow detection.
\newblock {\em Proceedings of Asian Conference on Computer Vision ACCV}, 15,
  2000.

\bibitem{hou2023survey}
B. Hou, Y. Liu, N. Ling, Y. Ren, L. Liu, et~al.
\newblock A survey of efficient deep learning models for moving object
  segmentation.
\newblock {\em APSIPA Transactions on Signal and Information Processing},
  12(1), 2023.

\bibitem{icsik2018swcd}
{\c{S}}. I{\c{s}}{\i}k, K. {\"O}zkan, S. G{\"u}nal, and {\"O}mer~N. Gerek.
\newblock {SWCD}: a sliding window and self-regulated learning-based background
  updating method for change detection in videos.
\newblock {\em Journal of Electronic Imaging}, 27(2):023002--023002, 2018.

\bibitem{jeeva2015survey}
S. Jeeva and M. Sivabalakrishnan.
\newblock Survey on background modeling and foreground detection for real time
  video surveillance.
\newblock {\em Procedia Computer Science}, 50:566--571, 2015.

\bibitem{kim2004background}
K. Kim, T.~H. Chalidabhongse, D. Harwood, and L. Davis.
\newblock Background modeling and subtraction by codebook construction.
\newblock 5:3061--3064, 2004.

\bibitem{lim2018foreground}
L.~A. Lim and H.~Y. Keles.
\newblock Foreground segmentation using convolutional neural networks for
  multiscale feature encoding.
\newblock {\em Pattern Recognition Letters}, 112:256--262, 2018.

\bibitem{lim2020learning}
L.~A. Lim and H.~Y. Keles.
\newblock Learning multi-scale features for foreground segmentation.
\newblock {\em Pattern Analysis and Applications}, 23(3):1369--1380, 2020.

\bibitem{maddalena2015towards}
L. Maddalena and A. Petrosino.
\newblock Towards benchmarking scene background initialization.
\newblock {\em New Trends in Image Analysis and Processing}, pages 469--476,
  2015.

\bibitem{mahadevan2009spatiotemporal}
V. Mahadevan and N. Vasconcelos.
\newblock Spatiotemporal saliency in dynamic scenes.
\newblock {\em IEEE Transactions on Pattern Analysis and Machine Intelligence},
  32(1):171--177, 2009.

\bibitem{oktay2018attention}
O. Oktay, J. Schlemper, L.~L. Folgoc, M. Lee, M. Heinrich, K. Misawa, K. Mori,
  S. McDonagh, N.~Y. Hammerla, B. Kainz, et~al.
\newblock Attention {U-N}et: {L}earning where to look for the pancreas.
\newblock {\em arXiv:1804.03999}, 2018.

\bibitem{radke2005image}
R.~J. Radke, S. Andra, O. Al-Kofahi, and B. Roysam.
\newblock Image change detection algorithms: a systematic survey.
\newblock {\em IEEE Transactions on Image Processing}, 14(3):294--307, 2005.

\bibitem{rahmon2021motion}
G. Rahmon, F. Bunyak, G. Seetharaman, and K. Palaniappan.
\newblock Motion {U-N}et: {M}ulti-cue encoder-decoder network for motion
  segmentation.
\newblock {\em 25th International Conference on Pattern Recognition}, pages
  8125--8132, 2021.

\bibitem{ronneberger2015u}
O. Ronneberger, P. Fischer, and T. Brox.
\newblock {U-N}et: Convolutional networks for biomedical image segmentation.
\newblock {\em 18th International Conference on Medical Image Computing and
  Computer-Assisted Intervention}, pages 234--241, 2015.

\bibitem{sakkos2018end}
D. Sakkos, H. Liu, J. Han, and L. Shao.
\newblock End-to-end video background subtraction with 3{D} convolutional
  neural networks.
\newblock {\em Multimedia Tools and Applications}, 77:23023--23041, 2018.

\bibitem{salehi2017tversky}
S.~S.~M. Salehi, D. Erdogmus, and A. Gholipour.
\newblock Tversky loss function for image segmentation using 3d fully
  convolutional deep networks.
\newblock {\em International Workshop on Machine Learning in Medical Imaging},
  pages 379--387, 2017.

\bibitem{sen2004robust}
S.~C. Sen-Ching and C. Kamath.
\newblock Robust techniques for background subtraction in urban traffic video.
\newblock {\em Visual Communications and Image Processing}, 5308:881--892,
  2004.

\bibitem{st2014subsense}
P.-L. St-Charles, G.-A. Bilodeau, and R. Bergevin.
\newblock Su{BSENSE}: {A} universal change detection method with local adaptive
  sensitivity.
\newblock {\em IEEE Transactions on Image Processing}, 24(1):359--373, 2014.

\bibitem{stauffer1999adaptive}
Chris Stauffer and W~Eric~L Grimson.
\newblock Adaptive background mixture models for real-time tracking.
\newblock {\em Proceedings of IEEE Computer Society Conference on Computer
  Vision and Pattern Recognition}, 2:246--252, 1999.

\bibitem{tezcan2019fully}
M.~O. Tezcan, J. Konrad, and P. Ishwar.
\newblock A fully-convolutional neural network for background subtraction of
  unseen videos.
\newblock {\em in the IEEE Proceedings of Winter Conf. Applications of Computer
  Vision}, pages 2774--2783, 2020.

\bibitem{vacavant2013benchmark}
A. Vacavant, T. Chateau, A. Wilhelm, and L. Lequievre.
\newblock A benchmark dataset for outdoor foreground/background extraction.
\newblock {\em Asian Conference on Computer Vision}, pages 291--300, 2013.

\bibitem{wang2014cdnet}
Y. Wang, P.-M. Jodoin, F. Porikli, J. Konrad, Y. Benezeth, and P. Ishwar.
\newblock {CD}net 2014: {A}n expanded change detection benchmark dataset.
\newblock {\em Proceedings of the IEEE Conference on Computer Vision and
  Pattern Recognition Workshops}, pages 387--394, 2014.

\bibitem{wang2017interactive}
Y. Wang, Z. Luo, and P.-M. Jodoin.
\newblock Interactive deep learning method for segmenting moving objects.
\newblock {\em Pattern Recognition Letters}, 96:66--75, 2017.

\bibitem{zeng2018background}
D. Zeng and M. Zhu.
\newblock Background subtraction using multiscale fully convolutional network.
\newblock {\em IEEE Access}, 6:16010--16021, 2018.

\bibitem{zeng2019combining}
D. Zeng, M. Zhu, and A. Kuijper.
\newblock Combining background subtraction algorithms with convolutional neural
  network.
\newblock {\em Journal of Electronic Imaging}, 28(1):013011--013011, 2019.

\bibitem{zheng2020novel}
W. Zheng, K. Wang, and F.-Y. Wang.
\newblock A novel background subtraction algorithm based on parallel vision and
  bayesian gans.
\newblock {\em Neurocomputing}, 394:178--200, 2020.

\bibitem{zhu2015human}
S. Zhu, L. Xia, et~al.
\newblock Human action recognition based on fusion features extraction of
  adaptive background subtraction and optical flow model.
\newblock {\em Mathematical Problems in Engineering}, 2015, 2015.

\end{thebibliography}
}

\end{document}